\documentclass[10pt, a4paper]{article}

\usepackage[final]{lrec2026} 

\usepackage{booktabs}
\usepackage{graphicx}
\usepackage{subcaption}
\usepackage{fancyvrb}
\usepackage{multirow}
\usepackage{amsmath}
\usepackage[table]{xcolor}

\usepackage{xcolor}
\newcommand{\hall}[1]{\textcolor{red}{#1}}

\title{A Critical Study of Automatic Evaluation in Sign Language Translation}

\name{Shakib Yazdani$^{1}$, Yasser Hamidullah$^{1}$, Cristina España-Bonet$^{1,2}$, \\
{\bf \large Eleftherios Avramidis$^{1}$, Josef van Genabith$^{1}$}}

\address{$^{1}$German Research Center for Artificial Intelligence (DFKI GmbH), \\
        Saarland Informatics Campus, Saarbrücken, Germany \\
        $^{2}$Barcelona Supercomputing Center (BSC-CNS), Barcelona, Catalonia, Spain \\
         \{shakib.yazdani,yasser.hamidullah,cristinae,eleftherios.avramidis,josef.van\_genabith\}@dfki.de\\}

\abstract{
Automatic evaluation metrics are crucial for advancing sign language translation (SLT). Current SLT evaluation metrics, such as BLEU and ROUGE, are only text-based, and it remains unclear to what extent text-based metrics can reliably capture the quality of SLT outputs. To address this gap, we investigate the limitations of text-based SLT evaluation metrics by analyzing six metrics, including BLEU, chrF, and ROUGE, as well as BLEURT on the one hand, and large language model (LLM)-based evaluators such as G-Eval and GEMBA zero-shot direct assessment on the other hand. Specifically, we assess the consistency and robustness of these metrics under three controlled conditions: paraphrasing, hallucinations in model outputs, and variations in sentence length. Our analysis highlights the limitations of lexical overlap metrics and demonstrates that while LLM-based evaluators better capture semantic equivalence often missed by conventional metrics, they can also exhibit bias toward LLM-paraphrased translations. Moreover, although all metrics are able to detect hallucinations, BLEU tends to be overly sensitive, whereas BLEURT and LLM-based evaluators are comparatively lenient toward subtle cases. This motivates the need for multimodal evaluation frameworks that extend beyond text-based metrics to enable a more holistic assessment of SLT outputs.
\\ \newline  }

\begin{document}

\maketitleabstract

\section{Introduction}

Sign languages are the primary communication systems for millions of deaf and hard-of-hearing people across the globe. They are highly expressive visual languages that convey meaning through hand signs, facial expressions, mouthings, and body posture \citep{stokoe1980sign}. Automatic sign language translation (SLT) aims to translate sign language videos into spoken language text to help bridge the communication gap between the deaf and hearing communities \citep{Camgoz_2018_CVPR}. However, evaluating SLT systems remains a significant challenge, since current evaluation practices rely solely on textual outputs, offering little insight into how effectively the visual-linguistic nature of sign language is captured. Human evaluation, though often the most reliable and serving as the gold standard in evaluating SLT, is costly and time-consuming, making it difficult to scale. As a result, current SLT models typically rely on text-based automatic evaluation metrics. Consequently, a valid measure of translation quality that accounts for this cross-modal grounding is still missing in SLT \citep{muller-etal-2023-findings}. In addition to the limitations of current evaluation metrics, SLT systems are also prone to hallucinations \citep{zhang2023sltunet, hamidullah2025grounding}, fluent but incorrect translations that misrepresent the signed input. Evaluating how metrics respond to such cases is crucial for understanding their reliability. 

When translating from sign language to spoken language text, traditional lexical overlap machine translation (MT) metrics such as BLEU \citep{papineni-etal-2002-bleu} and ROUGE \citep{lin-2004-rouge} are commonly adopted. However, these text-based metrics fall short in capturing the multimodal complexity of sign language.
To address this gap, recent studies have proposed evaluation metrics tailored to sign language tasks. SignBLEU \citep{kim-etal-2024-signbleu}, for instance, operates at the gloss level and aligns better with human judgments, while SiLVERScore \citep{imai-EtAl:2025:RANLP2} introduces an embedding-based approach for assessing sign language generation. Yet, SignBLEU depends on gloss annotations, and SiLVERScore has not been evaluated in the context of SLT.
Beyond these, embedding-based and large language model (LLM)-based metrics developed for text-to-text MT—such as BLEURT \citep{sellam-etal-2020-bleurt} and GEMBA \citep{kocmi-federmann-2023-gemba} have shown strong correlations with human judgments in text-to-text MT \citep{freitag-etal-2024-llms}. Nevertheless, these approaches remain underexplored in SLT, and there is still no comprehensive analysis of their strengths and limitations for evaluating SLT systems.

In this work, we systematically examine this evaluation gap through a comparative analysis of three lexical overlap metrics (BLEU, ROUGE, and chrF), one embedding-based metric (BLEURT), and two LLM-based evaluators (G-Eval and GEMBA), applied to four recent off-the-shelf SLT models, one gloss-based (TwoStream-SLT) and three gloss-free (SEM-SLT, SpaMo, and Signformer). The evaluation is performed under three controlled scenarios: paraphrasing (including word reordering), hallucinations in model outputs, and variations in sentence length. We summarize our contributions as follows: (1) we show that automatic lexical overlap metrics such as BLEU, chrF, and ROUGE are sensitive to surface-level lexical variation (paraphrasing) rather than true semantic equivalence. In contrast, embedding-based metric BLEURT and LLM-based evaluators G-Eval and GEMBA better reflect semantic similarity and meaning preservation in SLT outputs; (2) our analysis shows that all metrics are able to reliably distinguish hallucinated from non-hallucinated outputs. However, BLEU is highly sensitive in cases of extreme hallucination, whereas BLEURT and LLM-based evaluators are able to better capture meaning, though they tend to under-penalize subtle hallucinations when the translation remains fluent; (3) a fine-grained analysis by sentence length indicates systematic inconsistencies between metric types: lexical overlap metrics often rank models differently than BLEURT or LLM-based evaluators, highlighting that evaluation outcomes can vary depending on the chosen metric and sentence length.

\section{Related Work}
\label{sec:related_work}

\subsection{Evaluation Metrics}
SLT models have traditionally relied on lexical overlap MT metrics for automatic translation. Specifically, BLEU \citep{papineni-etal-2002-bleu} and ROUGE \citep{lin-2004-rouge} have been widely used following the neural SLT work by \citet{Camgoz_2018_CVPR}. Following \citet{muller-etal-2022-findings}, studies in SLT have adopted the embedding-based metric BLEURT \citep{sellam-etal-2020-bleurt}. However, these metrics do not take the visual source input into account. Recently, \citet{kim-etal-2024-signbleu} proposed SignBLEU, an evaluation metric designed to reduce information loss when assessing SLT outputs. By operating at the gloss level, it achieves stronger alignment with human judgments than prior text-based metrics. Nevertheless, SignBLEU remains constrained to the text modality and relies on manually curated gloss annotations, which limits scalability. More recently, \citet{imai-EtAl:2025:RANLP2} introduced SiLVERScore, an embedding-based metric that evaluates sign language generation in a multimodal semantic space. While it shows promising results, SiLVERScore was not evaluated on SLT.

The progress and advancements in LLMs have motivated research into their potential applications for automated assessment across various tasks. \citet{liu-etal-2023-g} pioneered LLM-based automated assessment by proposing \textsc{G-Eval}, a framework that uses Chain-of-Thought reasoning \citep{10.5555/3600270.3602070} along with task descriptions and evaluation metrics to evaluate performance on tasks including text summarization and dialogue generation. Following this direction, \citet{kocmi-federmann-2023-gemba} proposed \textsc{GEMBA}, a GPT-based evaluation metric to assess the MT quality both with- and without reference in a zero-shot prompting fashion. Recently, inspired by \textsc{G-Eval}, \citet{Tong_He_Shao_Yeung_2025} extended the framework to image and video caption evaluation using GPT-4o, achieving a high correlation with human judgments.

\subsection{Sign Language Translation}
SLT aims to convert sign language videos into spoken language text. SLT models are often categorized into gloss-free and methods with gloss supervision. \citet{Camgoz_2018_CVPR} pioneered SLT by framing it as a neural machine translation (NMT) task, and further extending it by leveraging transformers for end-to-end SLT \citep{camgoz2020sign}. MSKA-SLT \citep{10.1016/j.patcog.2025.111602} sets a new state-of-the-art on the Phoenix-2014T \citep{Camgoz_2018_CVPR} and CSL-Daily \citep{Zhou_2021_CVPR} benchmarks as a gloss-based model, with BLEU scores of $29.03$ and $25.52$, respectively. The approach models keypoints extracted from the face, body, and both hands using an attention mechanism, and is jointly optimized with a CTC loss and a self-distillation loss. However, glosses require manual human annotation, which is labor-intensive and difficult to scale. In contrast, recent approaches attempt to model SLT at the sentence level and a gloss-free fashion. Building on this objective, recent works have incorporated LLMs into SLT frameworks \citep{gong2024signllm, wong2024sign2gpt, chen-etal-2024-factorized}, employed contrastive learning objectives \citep{cheng2023cico}, replaced gloss supervision with sentence-level embeddings \citep{hamidullah2024slt}, and extended SLT to multilingual scenarios \citep{yin2022multislt, yazdani-etal-2025-continual, tan-etal-2025-multilingual, hamidullah-etal-2025-sonar}.

\subsection{Hallucination}

A model is said to hallucinate when it generates content that either lacks logical coherence or diverges from the source material, asserting information or events that are unfounded or inconsistent with the known facts \citep{survey-of-hallucination}. While hallucination detection has received increasing attention in the MT community, with studies on multilingual comparisons \citep{guerreiro-etal-2023-hallucinations} and LLM-based approaches for both low- and high-resource languages \citep{benkirane-etal-2024-machine}, research on hallucination in SLT remains limited. \citet{zhang2023sltunet} reported, based on manual analysis, that SLT models such as SLTUNET suffer greatly from hallucinations, where the generated translations often show limited correlation with the sign video. More recently, \citet{hamidullah2025grounding} proposed a hallucination detection framework and further demonstrated that gloss-free SLT models are particularly prone to severe hallucinations compared to their gloss-based counterparts. These findings highlight the need for a deeper understanding of how evaluation metrics respond to hallucinated outputs and how robust current SLT models are under such conditions.

\section{Experimental Setup}

\subsection{Task and Datasets}
In our experiments, we evaluate the limitations of current SLT metrics with respect to paraphrasing (including word reordering), hallucination sensitivity, and variation in translation length.
We use the Phoenix-2014T \footnote{Phoenix-2014T was selected because it is the most widely used benchmark in SLT research and contains gloss-level annotations necessary for evaluating gloss-based models.} dataset \citep{Camgoz_2018_CVPR}, covering weather forecasts in German sign language (DGS) to evaluate and compare the performance of various models in the above aspects.

\paragraph{Impact of paraphrasing on SLT evaluation.} 
For this aspect of our study, we compare recent off-the-shelf SLT models (Sec. \ref{sec:experiments:models}) across both traditional evaluation metrics and LLM-based ones (Sec. \ref{sec:experiments:metrics}) when we paraphrase the translation predicted by the model. We focus on paraphrasing, which naturally captures surface-level variations such as synonym substitution. We use \texttt{GPT-4o-mini} with the following prompt to generate paraphrased versions of model translations, which are then compared against their original references.

{\footnotesize
\begin{Verbatim}[commandchars=+\[\]]
Paraphrase the following sentence into a 
natural and fluent form. Do not alter any 
numbers written in words into digits or 
vice versa — keep the format as it is in 
the original text.
\end{Verbatim}
}

For lexical overlap metrics (BLEU, chrF, and ROUGE) and embedding-based BLEURT, and to ensure robustness against paraphrasing variations, we include a multi-reference evaluation in which the model’s translated and paraphrased output is compared against an augmented set of eleven references comprising the original human reference and ten gold-standard paraphrases.

\paragraph{Impact of hallucination on SLT evaluation.}

SLT models are prone to hallucinations, where the generated text diverges from the visual input. Yet, there has been no systematic investigation into how such hallucinations affect translation quality or the reliability of automatic SLT evaluation metrics. 
Motivated by findings from \citet{benkirane-etal-2024-machine}, which demonstrate the effectiveness of LLMs for hallucination detection in MT, we employ \texttt{Llama-3-70B} to identify hallucinated outputs in a reference-based setting.\footnote{The prompt used for hallucination detection is available in Appendix \ref{appendix:hallucination}.} To analyze how evaluation metrics behave under varying degrees of hallucination, we consider a \textbf{Severity Ranking} scheme that categorizes outputs into four levels: \textit{No Hallucination}, \textit{Small Hallucination}, \textit{Partial Hallucination}, and \textit{Full Hallucination}.

\paragraph{Impact of sentence length on SLT evaluation.}
In SLT, most studies report only the average evaluation score over the entire dataset, which can obscure important performance variations. 
To investigate the sensitivity of evaluation metrics to sentence length, we conduct a fine-grained analysis by segmenting the test set based on the number of words per sentence.
Specifically, we group sentences into five ranges: 1–6 (42 sentences), 7–12 (286 sentences), 13–18 (220 sentences), 19–24 (78 sentences), and 25–31 (16 sentences). This segmentation allows us to investigate how translation quality varies with sentence complexity and length, providing deeper insights into where existing models struggle and how hallucination patterns may correlate with sentence structure.

\begin{table*}[!htbp]
\centering
\resizebox{0.72\linewidth}{!}{%
\begin{tabular}{llcccc}
\toprule
Method & Setting & BLEU & chrF & BLEURT & ROUGE \\
\midrule
\multirow{3}{*}{Signformer} 
 & Original      & $14.8 \scriptstyle \pm 1.4$ & $35.0 \scriptstyle \pm 1.2$ & $0.425 \scriptstyle \pm 0.016$ & $32.8 \scriptstyle \pm 2.0$ \\[-1pt]
 & Paraphrased   & $5.7 \scriptstyle \pm 0.8$  & $31.7 \scriptstyle \pm 0.9$ & $0.452 \scriptstyle \pm 0.015$ & $27.7 \scriptstyle \pm 1.8$ \\[-1pt]
 & Multi-Ref     & $18.6 \scriptstyle \pm 1.4$      & $38.8 \scriptstyle \pm 1.3$      & $0.509 \scriptstyle \pm 0.016$         & $36.5 \scriptstyle \pm 1.8$ \\[2pt]
\cmidrule(lr){1-6}
\multirow{3}{*}{SEM-SLT} 
 & Original      & $23.7 \scriptstyle \pm 2.1$ & $45.8 \scriptstyle \pm 1.6$ & $0.484 \scriptstyle \pm 0.012$ & $47.9 \scriptstyle \pm 1.9$ \\[-1pt]
 & Paraphrased   & $10.0 \scriptstyle \pm 1.3$ & $38.3 \scriptstyle \pm 1.1$ & $0.537 \scriptstyle \pm 0.014$ & $38.0 \scriptstyle \pm 1.6$ \\[-1pt]
 & Multi-Ref     & $26.8 \scriptstyle \pm 2.4$ & $47.0 \scriptstyle \pm 1.7$ & $0.594 \scriptstyle \pm 0.013$ & $49.8 \scriptstyle \pm 1.8$ \\[2pt]
\cmidrule(lr){1-6}
\multirow{3}{*}{SpaMo} 
 & Original      & $22.2 \scriptstyle \pm 2.0$ & $44.2 \scriptstyle \pm 1.2$ & $0.542 \scriptstyle \pm 0.015$ & $43.2 \scriptstyle \pm 2.1$ \\[-1pt]
 & Paraphrased   & $9.3 \scriptstyle \pm 1.0$  & $38.6 \scriptstyle \pm 0.9$ & $0.558 \scriptstyle \pm 0.013$ & $35.3 \scriptstyle \pm 1.7$ \\[-1pt]
 & Multi-Ref     & $26.1 \scriptstyle \pm 1.8$ & $48.1 \scriptstyle \pm 1.1$ & $0.618 \scriptstyle \pm 0.015$ & $46.7 \scriptstyle \pm 1.3$ \\[2pt]
\cmidrule(lr){1-6}
\multirow{3}{*}{TwoStream-SLT} 
 & Original      & $28.2 \scriptstyle \pm 2.1$ & $50.5 \scriptstyle \pm 1.4$ & $0.597 \scriptstyle \pm 0.013$ & $50.3 \scriptstyle \pm 2.1$ \\[-1pt]
 & Paraphrased   & $12.7 \scriptstyle \pm 0.8$  & $43.3 \scriptstyle \pm 0.9$ & $0.604 \scriptstyle \pm 0.011$ & $40.5 \scriptstyle \pm 1.6$ \\[-1pt]
 & Multi-Ref     & $34.2 \scriptstyle \pm 1.6$  & $54.5 \scriptstyle \pm 1.1$ & $0.670 \scriptstyle \pm 0.009$ & $53.7 \scriptstyle \pm 1.4$ \\[2pt]
\bottomrule
\end{tabular}
}
\caption{Performance comparison of SLT models on the Phoenix-2014T test set under original, paraphrased, and multi-reference conditions across multiple evaluation metrics.}
\label{tab:results-compact}
\end{table*}

\subsection{Evaluation Metrics}
\label{sec:experiments:metrics}
For our analysis, we consider the metrics as per \citet{muller-etal-2022-findings,muller-etal-2023-findings}, specifically BLEU\footnote{\texttt{BLEU|nrefs:1|bs:25|tok:none|eff:no|
case:mixed|smooth:exp|version:2.5.1}} 
\citep[via SacreBLEU;][]{post-2018-call} for lexical overlap, 
ROUGE \citep{lin-2004-rouge}\footnote{\texttt{ROUGE|metrics:rougeL|nrefs:1|
stemmer:true|bs:25|version:0.1.2}} 
for recall-oriented n-gram overlap, chrF \citep{popovic-2015-chrf}\footnote{\texttt{chrF|nrefs:1|bs:25|case:mixed|eff:yes|
nc:6|nw:2|space:no|version:2.5.1}} that uses character $n$-grams, 
and embedding-based BLEURT \citep{sellam-etal-2020-bleurt}\footnote{BLEURT v0.0.2 using checkpoint BLEURT-20.} 
for semantic quality. We use bootstrap resampling for statistical significance reporting.

We also extend the traditional range of SLT evaluation metrics by presenting a sign language adapted version of G-Eval \citep{liu-etal-2023-g}. G-Eval introduces a framework that leverages prompting techniques to generate evaluation scores that strongly correlate with human preferences. The prompt is organized into three modules: (1) evaluation dimensions, which specify the aspects to be judged; (2) step-by-step reasoning, where Chain-of-Thought (CoT) \citep{10.5555/3600270.3602070} guides the LLM through the evaluation procedure; and (3) scoring with references, which constrains the output format and incorporates human translations as ground truth. Following \citet{sato-etal-2024-tmu}, we prompt the LLMs to evaluate the translation quality of the models in Section \ref{sec:experiments:models} for \textbf{Adequacy} and \textbf{Fluency}, each rated on a 5-point Likert scale ($1$–$5$).
Following \citet{fomicheva-etal-2022-mlqe, kocmi-federmann-2023-gemba}, we further include GEMBA zero-shot \textbf{Direct Assessment} \citep{kocmi-federmann-2023-gemba}, where the LLM is prompted to evaluate each translation hypothesis and provide a quality score ranging from $0$ (completely incorrect) to $100$ (perfect translation).
Our LLM experimental setup includes four models: \texttt{GPT-4.1-nano}, \texttt{Qwen3-8B}, \texttt{Llama-3.1-8B}, and \texttt{QWQ-32B}. These were chosen to include both a proprietary LLM and also the open-source ones.\footnote{All prompts used for G-Eval and GEMBA are provided in Appendices \ref{appendix:g-eval} and \ref{appendix:gemba}.}

\subsection{SLT Models}
\label{sec:experiments:models}
SLT models are either gloss-based or gloss-free. We include three gloss-free models: SpaMo \citep{hwang-etal-2025-spamo}, SEM-SLT \citep{hamidullah2024slt}, and Signformer \citep{yang2024signformerneededgeai}. We also include TwoStream-SLT \citep{10.5555/3600270.3601510} as a gloss-based model. We retrain these models to reproduce their reported results when the original model checkpoints are unavailable; however, our reproduced scores differ slightly from those originally reported.

\paragraph{SpaMo.} This model works by combining spatial and motion visual features using a multi-layer perceptron (MLP) and feeding them together with a prompt to an LLM for decoding and translation. Following the original setup, we set learning rate to $6\times10^{-4}$ and train for a maximum of 100 epochs.

\paragraph{SEM-SLT.} SLT models have traditionally relied on glosses for training. SEM-SLT avoids this reliance by supervising on sentence embeddings instead of glosses. The full pipeline sign2(sem+text) module, with an mBART decoder from SEM-SLT, was trained with a per-device batch size of 4 for both training and validation. The learning rate was set to $1\times10^{-5}$. Training was carried out with all other hyperparameters left at their default Hugging Face Trainer settings.

\paragraph{Signformer.} With the goal of edge AI, Signformer uses a combination of novel convolution, attention, and positional encoding to achieve competitive performance compared to gloss-based models while being quite smaller. Since the original paper did not specify how video features were extracted, we instead employ the S3D model pre-trained on both WLASL for ASL word recognition \citep{li2020word} and the Kinetics-400 dataset for human action recognition \citep{kay2017kineticshumanactionvideo}, following the approach by \citep{Chen2022ASM}. We set learning rate to 0.001 and train it for a maximum of 100 epochs.
 
\paragraph{TwoStream-SLT.} Unlike most SLT methods that extract visual embeddings from raw videos only, TwoStream-SLT leverages both key points (of hands, face, and upper body) and visual embeddings from raw videos to train a novel gloss-based network for sign language recognition and SLT. We used the pretrained Sign2Gloss and the Gloss2Text modules from the TwoStream-SLT repository to reproduce their results and outputs.

\begin{table*}[!htbp]
\centering
\small
\resizebox{0.8\linewidth}{!}{%
\begin{tabular}{lcccc}
\toprule
Method & Model & G-Eval Adequacy  & G-Eval Fluency & GEMBA \\
 & &  (Orig / Para) &  (Orig / Para) & ~~~~(Orig / Para)~~~~ \\
\midrule
Signformer     & gpt-4.1-nano & 2.54 / 2.57 & 3.34 / 3.48 & 38.45 / 38.33 \\
               & qwen3:8b     & 2.70 / 2.71 & 3.49 / 3.60 & 74.82 / 75.27 \\
               & qwq:32b      & 2.76 / 2.83 & 3.73 / 3.87 & 64.56 / 67.25 \\
               & llama3.1:8b  & 3.50 / 3.56 & 4.65 / 4.60 & 57.44 / 59.58 \\
\midrule
SEM-SLT        & gpt-4.1-nano & 2.76 / 2.91 & 3.26 / 3.60 & 47.53 / 53.17 \\
               & qwen3:8b     & 2.97 / 3.21 & 3.50 / 3.95 & 79.79 / 81.93 \\
               & qwq:32b      & 3.00 / 3.20 & 3.59 / 3.92 & 63.50 / 71.04 \\
               & llama3.1:8b  & 3.42 / 3.86 & 4.25 / 4.63 & 60.63 / 65.99 \\
\midrule
SpaMo          & gpt-4.1-nano & 3.00 / 3.06 & 3.66 / 3.73 & 55.36 / 57.31 \\
               & qwen3:8b     & 3.33 / 3.40 & 4.02 / 4.12 & 82.26 / 83.11 \\
               & qwq:32b      & 3.29 / 3.31 & 3.99 / 4.00 & 73.66 / 75.52 \\
               & llama3.1:8b  & 3.95 / 4.07 & 4.79 / 4.72 & 68.32 / 70.25 \\
\midrule
TwoStream-SLT  & gpt-4.1-nano & 3.28 / 3.33 & 3.77 / 3.82 & 64.63 / 66.69 \\
               & qwen3:8b     & 3.67 / 3.76 & 4.26 / 4.38 & 85.67 / 85.98 \\
               & qwq:32b      & 3.61 / 3.61 & 4.09 / 4.06 & 77.85 / 79.32 \\
               & llama3.1:8b  & 4.20 / 4.33 & 4.81 / 4.70 & 73.85 / 75.33 \\
\midrule
Signformer (Avg)     & - & 2.88 / 2.92 & 3.80 / 3.89 & 58.82 / 60.11 \\
SEM-SLT (Avg)        & - & 3.04 / 3.30 & 3.65 / 4.03 & 62.86 / 68.03 \\
SpaMo (Avg)          & - & 3.39 / 3.46 & 4.12 / 4.14 & 69.90 / 71.55 \\
TwoStream-SLT (Avg)  & - & 3.69 / 3.76 & 4.23 / 4.24 & 75.50 / 76.83 \\
\midrule
\end{tabular}
}
\caption{G-Eval adequacy and fluency scores, along with GEMBA, for original and paraphrased translations on the Phoenix-2014T test set.}
\label{tab:geval-merged}
\end{table*}

\section{Evaluation Results}

\subsection{Impact of Paraphrasing on SLT Evaluation}
One of the main limitations of lexical overlap metrics is their sensitivity to paraphrasing.
We provide a comparison of the scores with and without paraphrasing for every metric and model in Table~\ref{tab:results-compact}. Most prominently, we observe that lexical overlap metrics (BLEU, chrF, and ROUGE) are highly sensitive to variations in word order, even though our paraphrasing prompt preserves the overall sentence structure. Notably, the BLEU score drops to slightly less than half of its original value, while metrics such as chrF and ROUGE also decrease but to a much lesser extent. Interestingly, embedding-based BLEURT scores increase across all methods when comparing paraphrased translations to the original ones. This increase is particularly pronounced for SEM-SLT, increasing from $0.484$ to $0.537$. This is most likely because the SEM-SLT model is trained with a contrastive learning method using sentence-level embeddings, which allows it to generalize better to semantically equivalent paraphrases. To alleviate the sensitivity of lexical overlap metrics to paraphrasing and failure of capturing the overall semantics, we compare the metrics in a multi-reference setting, as shown in Table \ref{tab:results-compact}. We observe that including a multi-reference setting ensures that evaluation metrics fairly reward semantic equivalence when evaluating models that exhibit high surface-level variation, such as those subject to paraphrasing.

\begin{figure}[!h]
\centering
\includegraphics[width=\linewidth]{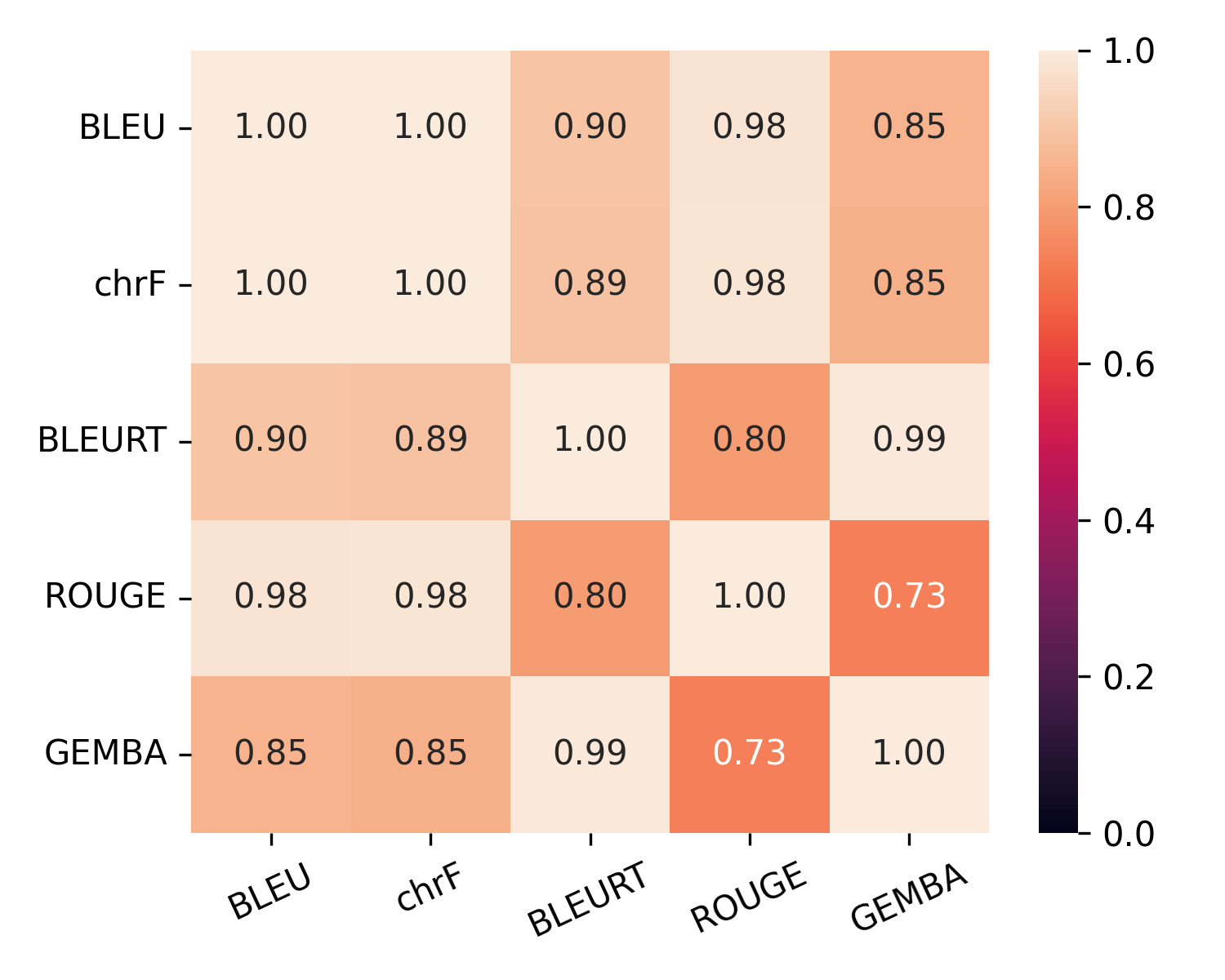}
    \caption{Pearson correlation between lexical metrics (BLEU, chrF, ROUGE), BLEURT, and GEMBA.}
    \label{fig:correlation_plot}
\end{figure}

We present the results of LLM-based evaluators, G-Eval (Adequacy and Fluency) and GEMBA, comparing paraphrased and original translations on the Phoenix-2014T test set in Table~\ref{tab:geval-merged}. Most notably, paraphrasing generally improves G-Eval Adequacy and Fluency and GEMBA scores, consistent with the pattern observed for BLEURT. \texttt{Llama3.1:8b} is the only model that rates paraphrased translations as less fluent than the original ones. Among all evaluated LLMs, \texttt{Llama3.1:8b} tends to assign the highest Adequacy and Fluency scores, whereas \texttt{gpt-4.1-nano} assigns the lowest. We observe that LLM-based evaluators tend to prefer paraphrased translations over the original ones. This tendency likely arises because the paraphrases are themselves generated by an LLM, which may introduce stylistic patterns or linguistic preferences that align with the evaluator’s own training distribution, thereby biasing its judgments. Interestingly, this observation aligns with previous studies suggesting that LLMs exhibit a bias toward LLM-generated or paraphrased text~\citep{xu-etal-2024-pride, liu-etal-2023-g}. 
To be able to make more general statements about evaluation metrics, we show the Pearson correlation between the lexical metrics, BLEURT, and GEMBA, computed using the average scores from Table \ref{tab:geval-merged} in Figure \ref{fig:correlation_plot}. We notice a high correlation among lexical overlap metrics, and GEMBA is also highly correlated with BLEURT, as both incorporate learned language model representations in their scoring. Our previous observations from Tables \ref{tab:results-compact} and \ref{tab:geval-merged} are confirmed, with lexical metrics (BLEU, chrF, ROUGE) yielding consistent rankings, while LLM-based evaluations align closely with BLEURT, ranking TwoStream-SLT highest, followed by SpaMo, SEM-SLT, and Signformer.

\begin{figure*}[!htbp]
\centering
\includegraphics[width=\linewidth]{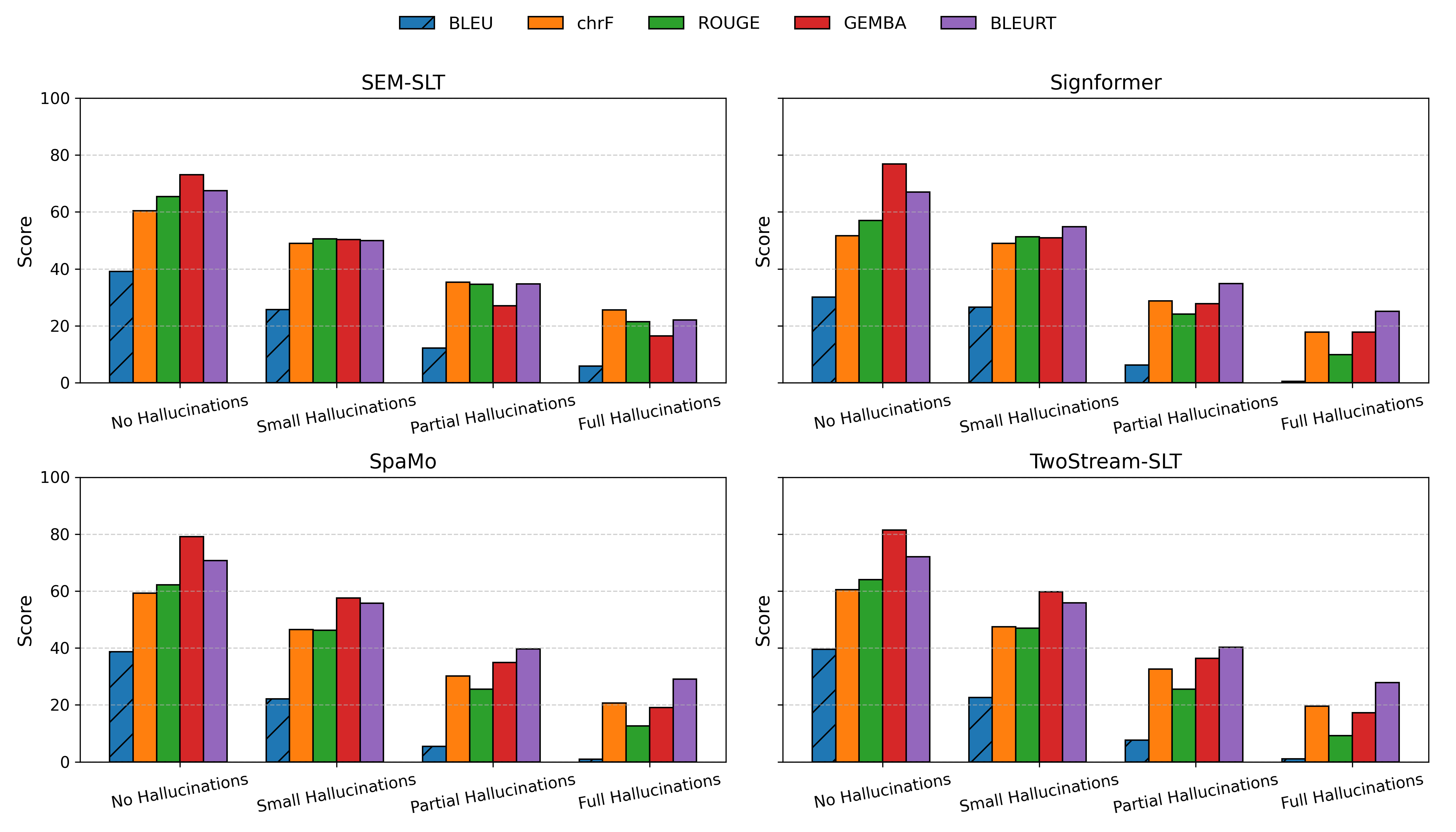}
    \caption{Examination of the sensitivity of evaluation metrics, including BLEU, chrF, ROUGE, GEMBA, and BLEURT, to hallucinations in SLT outputs on the Phoenix-2014T test set.}
    \label{fig:combined_bar_plots_hallu}
\end{figure*}

\subsection{Impact of Hallucination on SLT Evaluation}

We first analyze how hallucinations impact automatic metrics and then examine how robust SLT models are in the case of hallucination. Figure~\ref{fig:combined_bar_plots_hallu} illustrates the impact of hallucinations, categorized by their severity ranking, on various evaluation metrics across four SLT models. We omitted G-Eval in this setting because its results are reported on a Likert scale. \footnote{Typically ranging from 1 (strongly disagree/poor) to 5 (strongly agree/excellent) \citep{likert1932technique}.} Moreover, we report hallucination detection results for GEMBA using \texttt{gpt-4.1-nano}. Notably, results show that there is a monotonic decline across all metrics, meaning that as hallucination severity level increases, translation quality drops. However, BLEU is an extremely sensitive metric; in cases such as \textit{Partial Hallucinations} or \textit{Full Hallucinations}, even minor rephrasings that remain semantically correct can cause a sharp decrease in BLEU scores. ROUGE and chrF decline more smoothly, since they account for partial overlaps at the subword or character level, respectively. GEMBA and BLEURT follow a similar trend, as they can still capture some semantic meaning even in extreme \textit{Full Hallucination} cases where the surface wording differs substantially. However, we suspect that these metrics may under-penalize subtle hallucinations when the generated text is fluent, thereby overestimating the overall score. Similar observations regarding BLEURT’s overestimation and the data leakage issues in LLMs have been discussed by \citet{zeng-etal-2024-towards}.

\begin{figure*}[!htbp]
\centering
\includegraphics[width=\linewidth]{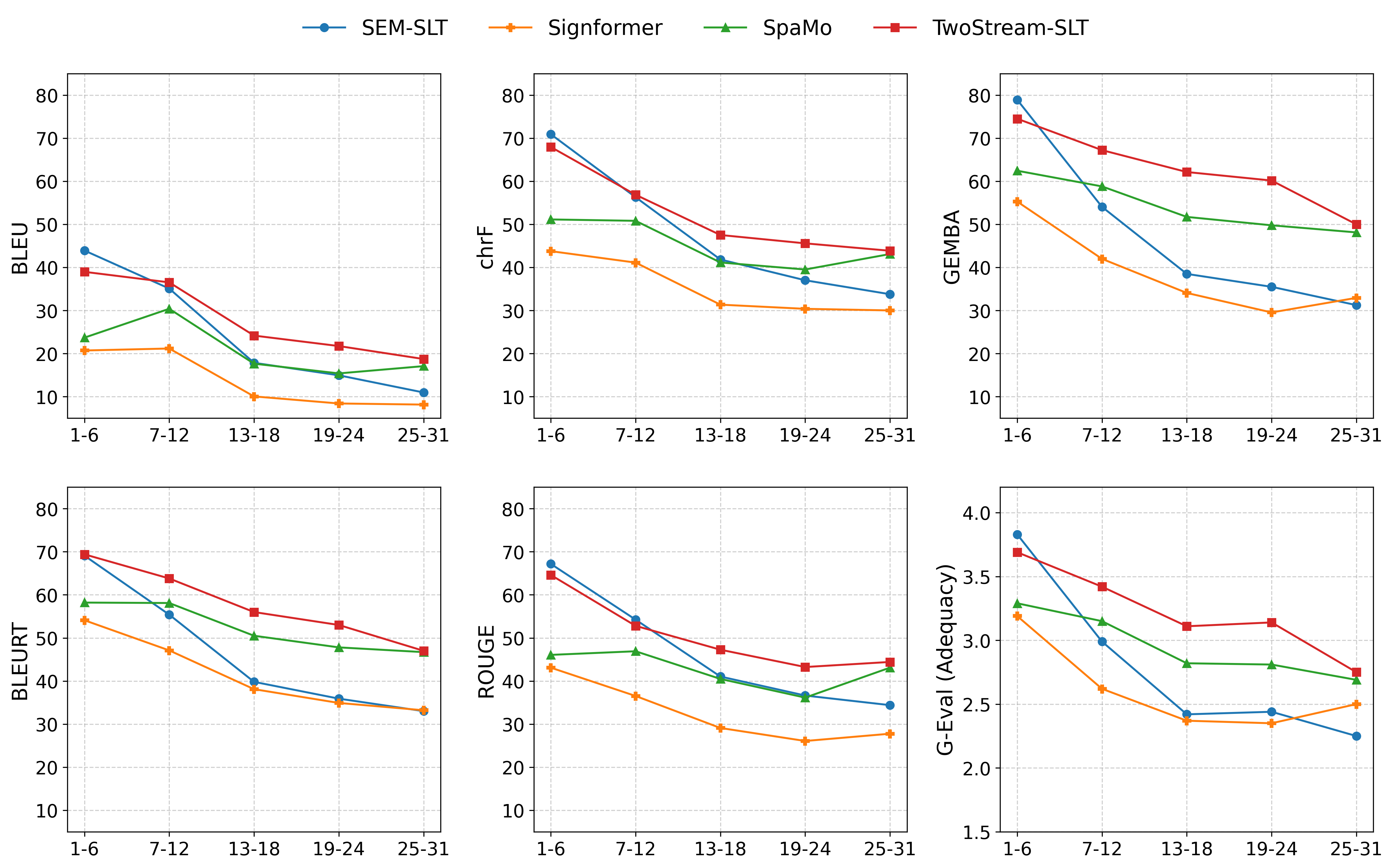}
    \caption{Evaluation of four SLT models (SEM-SLT, Signformer, SpaMo, TwoStream-SLT) across sentence length bins (1–6, 7–12, 13–18, 19–24, 25–31) on the Phoenix-2014T test set.}
    \label{fig:combined_line_plots_trans}
\end{figure*}

Different from the main focus of the paper that concentrates on evaluating evaluation metrics, here we briefly take the opportunity to investigate the behavior of SLT models when hallucinations occurs. Overall, TwoStream-SLT and SpaMo demonstrate a consistent trend across all evaluation metrics, showing less hallucinations compared to other models. In contrast, Signformer exhibits notably lower average performance, aligning with the results presented in Table \ref{tab:results-compact}. To analyze the extent to which SLT models produce hallucinations, we present in Table \ref{tab:hallucination_counts} a breakdown of the number of hallucinated translations according to severity level. The results reveal that gloss-based TwoStream-SLT yields the lowest \textit{Full Hallucinations} (24), indicating higher robustness to severe errors, whereas Signformer produces the highest number (177), confirming our earlier observations regarding its overall vulnerability to hallucination. This finding is in line with previous work showing that gloss-free models exhibit higher hallucination rates than gloss-based models \citep{hamidullah2025grounding}.

\begin{table}[!htbp]
\centering
\resizebox{\linewidth}{!}{
\begin{tabular}{lcccc}
\toprule
Model & No & Small & Partial & Full \\
\midrule
Signformer   & 87  & 193 & 185 & 177 \\
SEM-SLT          & 187 & 222 & 174 & 59  \\
SpaMo            & 204 & 231 & 138 & 69  \\
TwoStream-SLT    & 296 & 228 & 94  & 24  \\
\bottomrule
\end{tabular}
}
\caption{Number of hallucinations based on severity level (No, Small, Partial, and Full Hallucinations) across SLT models.}
\label{tab:hallucination_counts}
\end{table}

\subsection{Impact of Sentence Length on SLT Evaluation}

To highlight the limitations of text-based automatic metrics in SLT, we evaluate their performance across five sentence-length bins, as shown in Figure \ref{fig:combined_line_plots_trans}. We omitted G-Eval (Fluency) for this experiment as we are concerned with the overall translation quality. Across all evaluation metrics, performance consistently decreases with sentence length, indicating that longer sentences pose greater challenges for current SLT models, though the rate and shape of decline vary by metric. Notably, we observe that SEM-SLT ranks third according to BLEURT or LLM-based (GEMBA and G-Eval) metrics but first or second under lexical overlap ones, revealing a metric inconsistency.

Additionally, we analyze how translation quality varies with sentence length across different SLT models. Notably, gloss-based TwoStream-SLT demonstrates the strongest overall performance and exhibits the least sensitivity to longer sequences. SEM-SLT performs competitively on short sentences (1–6 tokens) but its performance declines more sharply for longer sentences, particularly on BLEU, GEMBA, and G-Eval. SpaMo shows moderate yet consistent performance, with relatively smaller drops for longer sequences compared to the other models. Finally, Signformer underperforms across all metrics and sentence-length bins, consistent with the trends observed in our earlier experiments.

\section{Qualitative Analysis} 

To better understand the impact of paraphrasing and hallucination on translation quality, we analyze specific examples exhibiting extreme cases of hallucination. Specifically, \textit{Partial Hallucinations} and \textit{Full Hallucinations}, where the GEMBA score exceeds 50. Notably, we did not find any instances of \textit{Full Hallucinations} with a GEMBA score above 50 across all SLT models. Table \ref{tab:qual_partial_hallucination_colored} presents one illustrative example per SLT model. We observe that paraphrasing tends to increase the GEMBA score in cases of \textit{Partial Hallucinations} across all models. By examining the paraphrased examples, we find that paraphrased translations are generally more fluent and natural than the original outputs, even though hallucinations persist. However, this is not always the case, as in most examples, paraphrasing either leaves the GEMBA score unchanged or leads to a decrease.

\begin{table}[!h]
\centering
\small
\renewcommand{\arraystretch}{0.9}
\setlength{\tabcolsep}{2.5pt}
\begin{tabular}{p{1.2 cm} p{6.1cm}}
\toprule
Model & SpaMo \\
\addlinespace[0.7ex]
GEMBA & 65 $\rightarrow$ 85 \\
\addlinespace[0.7ex]
Ref. & am wochenende beruhigt sich dann das wetter langsam . \\
\addlinespace[0.7ex]
& \textcolor{gray}{(Over the weekend, the weather gradually calms down.)} \\
\addlinespace[0.7ex]
Pred. & am wochenende erwartet uns dann \hall{verbreitet} ruhiges \hall{recht trockenes} \hall{winterwetter} . \\
\addlinespace[0.7ex]
Paraph. & am wochenende erwartet uns \hall{weitgehend} ruhiges und \hall{recht trockenes} \hall{winterwetter} . \\
\midrule
Model & SEM-SLT \\
\addlinespace[0.7ex]
GEMBA & 65 $\rightarrow$ 85 \\
\addlinespace[0.7ex]
Ref. & im norden und nordosten bleibt es meist bedeckt mitunter fällt dort etwas regen . \\
\addlinespace[0.7ex]
& \textcolor{gray}{(In the north and northeast, it mostly stays cloudy, with occasional light rain there.)} \\
\addlinespace[0.7ex]
Pred. & im norden und nordosten fällt regen \hall{sonst ist es meist trocken} . \\
\addlinespace[0.7ex]
Paraph. & im norden und nordosten regnet es, \hall{während es andernorts meist trocken bleibt} . \\
\midrule
Model & TwoStream-SLT \\
\addlinespace[0.7ex]
GEMBA & 75 $\rightarrow$ 85 \\
\addlinespace[0.7ex]
Ref. & in der südhälfte muss dazu mit nachtfrost gerechnet werden . \\
\addlinespace[0.7ex]
& \textcolor{gray}{(In the southern half, one must expect nighttime frost.)} \\
\addlinespace[0.7ex]
Pred. & im süden \hall{gibt es heute nacht} teilweise \hall{strengen} frost . \\
\addlinespace[0.7ex]
Paraph. & im süden wird es \hall{heute nacht} teilweise \hall{sehr} frostig sein . \\
\midrule
Model & Signformer \\
\addlinespace[0.7ex]
GEMBA & 75 $\rightarrow$ 85 \\
\addlinespace[0.7ex]
Ref. & südlich des mains morgen verbreitet freundlich . \\
\addlinespace[0.7ex]
& \textcolor{gray}{(South of the Main, it will be generally sunny tomorrow.)} \\
\addlinespace[0.7ex]
Pred. & \hall{vor allem in der südwesthälfte} ist es \hall{länger} freundlich . \\
\addlinespace[0.7ex]
Paraph. & \hall{insbesondere in der südwestlichen hälfte} bleibt es \hall{länger} freundlich . \\
\bottomrule
\end{tabular}
\caption{Qualitative examples in the presence of \textit{Partial Hallucinations} with paraphrased predictions.}
\label{tab:qual_partial_hallucination_colored}
\end{table}

\section{Conclusion}

In this work, we present a systematic analysis of text-based automatic evaluation metrics for SLT, including traditional lexical overlap metrics such as BLEU and ROUGE, and recent embedding- and LLM-based metrics such as BLEURT, GEMBA, and G-Eval. Our results highlight the limitations of both categories: lexical metrics (BLEU, chrF, ROUGE) often fail to capture semantic adequacy, particularly under paraphrasing, whereas embedding-based BLEURT and LLM-based evaluators (G-Eval and GEMBA), despite being more semantically aware, tend to overestimate the quality of fluent but hallucinated outputs. Based on these findings, we recommend using BLEURT or LLM-based evaluators such as GEMBA and G-Eval along with traditional lexical overlap metrics. Furthermore, we advocate for the development and adoption of multimodal evaluation frameworks that extend beyond text-based metrics to provide a more holistic and comprehensive assessment of SLT outputs.

\section*{Limitations}

In this work, we investigate the limitations of text-based automatic evaluation metrics in SLT through empirical experiments. While our study covers large and controlled settings, it has several limitations. First, our evaluation relies solely on automatic metrics, as no human evaluation was conducted to serve as a gold standard. This limitation stems from the lack of available native signers for large-scale annotation. Future work should include human judgments from native sign language users to establish a more reliable evaluation benchmark. Second, our experiments are restricted to the Phoenix-2014T dataset, which contains German Sign Language (DGS) data primarily focused on weather forecasts. This narrow domain may limit the generalizability of our findings to broader SLT contexts. Finally, although our results consistently highlight the weaknesses of lexical overlap metrics, we rely primarily on BLEU for quantitative evaluation. However, BLEU scores can be unreliable when computed on small test sets: prior work has shown that BLEU’s correlation with human judgments becomes unstable when sample sizes are low \citep{mathur-etal-2020-tangled}. To mitigate this issue, we applied bootstrap resampling to estimate confidence intervals, which provides a more robust assessment of metric stability and significance. We emphasize that these experiments are part of an ongoing research effort, and the findings reported here are preliminary and should not be deployed or regarded as final without community approval.

\nocite{*}
\section{Bibliographical References}
\label{sec:reference}

\bibliographystyle{lrec2026-natbib}
\bibliography{lrec2026-example}

@InProceedings{Camgoz_2018_CVPR,
 author={Camgoz, Necati Cihan and Hadfield, Simon and Koller, Oscar and Ney, Hermann and Bowden, Richard},
  booktitle={2018 IEEE/CVF Conference on Computer Vision and Pattern Recognition}, 
  title={Neural Sign Language Translation}, 
  year={2018},
  volume={},
  number={},
  pages={7784-7793},
  doi={10.1109/CVPR.2018.00812}}

@inproceedings{muller-etal-2022-findings,
  author    = {Müller, Mathias  and  Ebling, Sarah  and  Avramidis, Eleftherios  and  Battisti, Alessia  and  Berger, Michèle  and  Bowden, Richard  and  Braffort, Annelies  and  Cihan Camgöz, Necati  and  España-Bonet, Cristina  and  Grundkiewicz, Roman  and  Jiang, Zifan  and  Koller, Oscar  and  Moryossef, Amit  and  Perrollaz, Regula  and  Reinhard, Sabine  and  Rios, Annette  and  Shterionov, Dimitar  and  Sidler-Miserez, Sandra  and  Tissi, Katja  and  Van Landuyt, Davy},
  title     = "{Findings of the First WMT Shared Task on Sign Language Translation (WMT-SLT22)}",
  booktitle      = {Proceedings of the Seventh Conference on Machine Translation},
  month          = {December},
  year           = {2022},
  address        = {Abu Dhabi},
  publisher      = {Association for Computational Linguistics},
  pages     = {744--772},
  url       = {https://aclanthology.org/2022.wmt-1.71}
}

@inproceedings{muller-etal-2023-findings,
    title = "Findings of the Second {WMT} Shared Task on Sign Language Translation ({WMT}-{SLT}23)",
    author = {M{\"u}ller, Mathias  and
      Alikhani, Malihe  and
      Avramidis, Eleftherios  and
      Bowden, Richard  and
      Braffort, Annelies  and
      Cihan Camg{\"o}z, Necati  and
      Ebling, Sarah  and
      Espa{\~n}a-Bonet, Cristina  and
      G{\"o}hring, Anne  and
      Grundkiewicz, Roman  and
      Inan, Mert  and
      Jiang, Zifan  and
      Koller, Oscar  and
      Moryossef, Amit  and
      Rios, Annette  and
      Shterionov, Dimitar  and
      Sidler-Miserez, Sandra  and
      Tissi, Katja  and
      Van Landuyt, Davy},
    editor = "Koehn, Philipp  and
      Haddow, Barry  and
      Kocmi, Tom  and
      Monz, Christof",
    booktitle = "Proceedings of the Eighth Conference on Machine Translation",
    month = dec,
    year = "2023",
    address = "Singapore",
    publisher = "Association for Computational Linguistics",
    url = "https://aclanthology.org/2023.wmt-1.4",
    doi = "10.18653/v1/2023.wmt-1.4",
    pages = "68--94",
}

@inproceedings{post-2018-call,
    title = "A Call for Clarity in Reporting {BLEU} Scores",
    author = "Post, Matt",
    booktitle = "Proceedings of the Third Conference on Machine Translation: Research Papers",
    month = oct,
    year = "2018",
    address = "Brussels, Belgium",
    publisher = "Association for Computational Linguistics",
    url = "https://aclanthology.org/W18-6319",
    doi = "10.18653/v1/W18-6319",
    pages = "186--191",
}

@inproceedings{lin-2004-rouge,
    title = "{ROUGE}: A Package for Automatic Evaluation of Summaries",
    author = "Lin, Chin-Yew",
    booktitle = "Text Summarization Branches Out",
    month = jul,
    year = "2004",
    address = "Barcelona, Spain",
    publisher = "Association for Computational Linguistics",
    url = "https://aclanthology.org/W04-1013/",
    pages = "74--81"
}

@inproceedings{sellam-etal-2020-bleurt,
    title = "{BLEURT}: Learning Robust Metrics for Text Generation",
    author = "Sellam, Thibault  and
      Das, Dipanjan  and
      Parikh, Ankur",
    editor = "Jurafsky, Dan  and
      Chai, Joyce  and
      Schluter, Natalie  and
      Tetreault, Joel",
    booktitle = "Proceedings of the 58th Annual Meeting of the Association for Computational Linguistics",
    month = jul,
    year = "2020",
    address = "Online",
    publisher = "Association for Computational Linguistics",
    url = "https://aclanthology.org/2020.acl-main.704/",
    doi = "10.18653/v1/2020.acl-main.704",
    pages = "7881--7892"
}

@inproceedings{popovic-2015-chrf,
    title = "chr{F}: character n-gram {F}-score for automatic {MT} evaluation",
    author = "Popovi{\'c}, Maja",
    booktitle = "Proceedings of the Tenth Workshop on Statistical Machine Translation",
    month = sep,
    year = "2015",
    publisher = "Association for Computational Linguistics",
    doi = "10.18653/v1/W15-3049",
    pages = "392--395",
}

@inproceedings{liu-etal-2023-g,
    title = "{G}-Eval: {NLG} Evaluation using Gpt-4 with Better Human Alignment",
    author = "Liu, Yang  and
      Iter, Dan  and
      Xu, Yichong  and
      Wang, Shuohang  and
      Xu, Ruochen  and
      Zhu, Chenguang",
    editor = "Bouamor, Houda  and
      Pino, Juan  and
      Bali, Kalika",
    booktitle = "Proceedings of the 2023 Conference on Empirical Methods in Natural Language Processing",
    month = dec,
    year = "2023",
    address = "Singapore",
    publisher = "Association for Computational Linguistics",
    url = "https://aclanthology.org/2023.emnlp-main.153/",
    doi = "10.18653/v1/2023.emnlp-main.153",
    pages = "2511--2522"
}

@article{openai2024gpt4ocard,
      title={GPT-4o System Card}, 
      author={OpenAI et al.},
      year={2024},
      eprint={2410.21276},
      archivePrefix={arXiv},
      primaryClass={cs.CL},
      url={https://arxiv.org/abs/2410.21276}, 
      journal={ArXiv}
}

@inproceedings{10.5555/3600270.3602070,
author = {Wei, Jason and Wang, Xuezhi and Schuurmans, Dale and Bosma, Maarten and Ichter, Brian and Xia, Fei and Chi, Ed H. and Le, Quoc V. and Zhou, Denny},
title = {Chain-of-thought prompting elicits reasoning in large language models},
year = {2022},
isbn = {9781713871088},
publisher = {Curran Associates Inc.},
address = {Red Hook, NY, USA},
booktitle = {Proceedings of the 36th International Conference on Neural Information Processing Systems},
articleno = {1800},
numpages = {14},
location = {New Orleans, LA, USA},
series = {NIPS '22}
}

@inproceedings{hamidullah2024slt,
    title = "Sign Language Translation with Sentence Embedding Supervision",
    author = "Hamidullah, Yasser  and
      van Genabith, Josef  and
      Espa{\~n}a-Bonet, Cristina",
    editor = "Ku, Lun-Wei  and
      Martins, Andre  and
      Srikumar, Vivek",
    booktitle = "Proceedings of the 62nd Annual Meeting of the Association for Computational Linguistics (Volume 2: Short Papers)",
    month = aug,
    year = "2024",
    address = "Bangkok, Thailand",
    publisher = "Association for Computational Linguistics",
    url = "https://aclanthology.org/2024.acl-short.40/",
    doi = "10.18653/v1/2024.acl-short.40",
    pages = "425--434"
}

@inproceedings{hwang-etal-2025-spamo,
    title = "An Efficient Gloss-Free Sign Language Translation Using Spatial Configurations and Motion Dynamics with {LLM}s",
    author = "Hwang, Eui Jun  and
      Cho, Sukmin  and
      Lee, Junmyeong  and
      Park, Jong C.",
    editor = "Chiruzzo, Luis  and
      Ritter, Alan  and
      Wang, Lu",
    booktitle = "Proceedings of the 2025 Conference of the Nations of the Americas Chapter of the Association for Computational Linguistics: Human Language Technologies (Volume 1: Long Papers)",
    month = apr,
    year = "2025",
    address = "Albuquerque, New Mexico",
    publisher = "Association for Computational Linguistics",
    url = "https://aclanthology.org/2025.naacl-long.197/",
    doi = "10.18653/v1/2025.naacl-long.197",
    pages = "3901--3920",
    ISBN = "979-8-89176-189-6"
}

@article{yang2024signformerneededgeai,
      title={Signformer is all you need: Towards Edge AI for Sign Language}, 
      author={Eta Yang},
      year={2024},
      eprint={2411.12901},
      archivePrefix={arXiv},
      primaryClass={cs.CL},
      url={https://arxiv.org/abs/2411.12901}, 
      journal={ArXiv}
}

@inproceedings{10.5555/3600270.3601510,
author = {Chen, Yutong and Zuo, Ronglai and Wei, Fangyun and Wu, Yu and Liu, Shujie and Mak, Brian},
title = {Two-stream network for sign language recognition and translation},
year = {2022},
isbn = {9781713871088},
publisher = {Curran Associates Inc.},
address = {Red Hook, NY, USA},
booktitle = {Proceedings of the 36th International Conference on Neural Information Processing Systems},
articleno = {1240},
numpages = {14},
location = {New Orleans, LA, USA},
series = {NIPS '22}
}

@inproceedings{sato-etal-2024-tmu,
    title = "{TMU}-{HIT}{'}s Submission for the {WMT}24 Quality Estimation Shared Task: Is {GPT}-4 a Good Evaluator for Machine Translation?",
    author = "Sato, Ayako  and
      Nakajima, Kyotaro  and
      Kim, Hwichan  and
      Chen, Zhousi  and
      Komachi, Mamoru",
    editor = "Haddow, Barry  and
      Kocmi, Tom  and
      Koehn, Philipp  and
      Monz, Christof",
    booktitle = "Proceedings of the Ninth Conference on Machine Translation",
    month = nov,
    year = "2024",
    address = "Miami, Florida, USA",
    publisher = "Association for Computational Linguistics",
    url = "https://aclanthology.org/2024.wmt-1.38/",
    doi = "10.18653/v1/2024.wmt-1.38",
    pages = "529--534"
}

@inproceedings{camgoz2020sign,
  author = {Necati Cihan Camgoz and Oscar Koller and Simon Hadfield and Richard Bowden},
  title = {Sign Language Transformers: Joint End-to-end Sign Language Recognition and Translation},
  booktitle = {IEEE Conference on Computer Vision and Pattern Recognition (CVPR)},
  year = {2020}
}

@inproceedings{
zhang2023sltunet,
title={{SLTUNET}: A Simple Unified Model for Sign Language Translation},
author={Biao Zhang and Mathias M{\"u}ller and Rico Sennrich},
booktitle={The Eleventh International Conference on Learning Representations },
year={2023},
url={https://openreview.net/forum?id=EBS4C77p_5S}
}

@article{stokoe1980sign,
  title={Sign language structure},
  author={Stokoe, William C.},
  journal={Annual Review of Anthropology},
  volume={9},
  pages={365--390},
  year={1980},
  publisher={Annual Reviews},
  doi={10.1146/annurev.an.09.100180.002053}
}

@article{10.1016/j.patcog.2025.111602,
author = {Guan, Mo and Wang, Yan and Ma, Guangkun and Liu, Jiarui and Sun, Mingzu},
title = {MSKA: Multi-stream keypoint attention network for sign language recognition and translation},
year = {2025},
issue_date = {Sep 2025},
publisher = {Elsevier Science Inc.},
address = {USA},
volume = {165},
number = {C},
issn = {0031-3203},
url = {https://doi.org/10.1016/j.patcog.2025.111602},
doi = {10.1016/j.patcog.2025.111602},
journal = {Pattern Recogn.},
month = may,
numpages = {10}
}

@InProceedings{Zhou_2021_CVPR,
    author    = {Zhou, Hao and Zhou, Wengang and Qi, Weizhen and Pu, Junfu and Li, Houqiang},
    title     = {Improving Sign Language Translation With Monolingual Data by Sign Back-Translation},
    booktitle = {Proceedings of the IEEE/CVF Conference on Computer Vision and Pattern Recognition (CVPR)},
    month     = {June},
    year      = {2021},
    pages     = {1316-1325}
}

@InProceedings{Yin_2023_CVPR,
    author    = {Yin, Aoxiong and Zhong, Tianyun and Tang, Li and Jin, Weike and Jin, Tao and Zhao, Zhou},
    title     = {Gloss Attention for Gloss-Free Sign Language Translation},
    booktitle = {Proceedings of the IEEE/CVF Conference on Computer Vision and Pattern Recognition (CVPR)},
    month     = {June},
    year      = {2023},
    pages     = {2551-2562}
}

@inproceedings{wong2024sign2gpt,
title={Sign2{GPT}: Leveraging Large Language Models for Gloss-Free Sign Language Translation},
author={Ryan Wong and Necati Cihan Camgoz and Richard Bowden},
booktitle={The Twelfth International Conference on Learning Representations},
year={2024},
url={https://openreview.net/forum?id=LqaEEs3UxU}
}

@inproceedings{chen-etal-2024-factorized,
    title = "Factorized Learning Assisted with Large Language Model for Gloss-free Sign Language Translation",
    author = "Chen, Zhigang  and
      Zhou, Benjia  and
      Li, Jun  and
      Wan, Jun  and
      Lei, Zhen  and
      Jiang, Ning  and
      Lu, Quan  and
      Zhao, Guoqing",
    booktitle = "Proceedings of the 2024 Joint International Conference on Computational Linguistics, Language Resources and Evaluation (LREC-COLING 2024)",
    month = may,
    year = "2024",
    address = "Torino, Italia",
    publisher = "ELRA and ICCL",
    url = "https://aclanthology.org/2024.lrec-main.620/",
    pages = "7071--7081",
}

@inproceedings{gong2024signllm,
    author    = {Gong, Jia and Foo, Lin Geng and He, Yixuan and Rahmani, Hossein and Liu, Jun},
    title     = {LLMs are Good Sign Language Translators},
    booktitle = {Proceedings of the IEEE/CVF Conference on Computer Vision and Pattern Recognition (CVPR)},
    month     = {June},
    year      = {2024},
    pages     = {18362-18372}
}

@inproceedings{cheng2023cico,
  title={CiCo: Domain-Aware Sign Language Retrieval via Cross-Lingual Contrastive Learning},
  author={Cheng, Yiting and Wei, Fangyun and Jianmin, Bao and Chen, Dong and Zhang, Wen Qiang},
  booktitle={CVPR},
  year={2023}
}

@inproceedings{tan-etal-2025-multilingual,
    title = "Multilingual Gloss-free Sign Language Translation: Towards Building a Sign Language Foundation Model",
    author = "Tan, Sihan  and
      Miyazaki, Taro  and
      Nakadai, Kazuhiro",
    booktitle = "Proceedings of the 63rd Annual Meeting of the Association for Computational Linguistics (Volume 2: Short Papers)",
    month = jul,
    year = "2025",
    address = "Vienna, Austria",
    publisher = "Association for Computational Linguistics",
    url = "https://aclanthology.org/2025.acl-short.43/",
    doi = "10.18653/v1/2025.acl-short.43",
    pages = "553--561",
    ISBN = "979-8-89176-252-7",
}

@inproceedings{yin2022multislt,
  author={Yin, Aoxiong and Zhao, Zhou and Jin, Weike and Zhang, Meng and Zeng, Xingshan and He, Xiaofei},
  booktitle={2022 IEEE/CVF Conference on Computer Vision and Pattern Recognition (CVPR)}, 
  title={MLSLT: Towards Multilingual Sign Language Translation}, 
  year={2022},
  volume={},
  number={},
  pages={5099-5109},
  doi={10.1109/CVPR52688.2022.00505}}

@inproceedings{yazdani-etal-2025-continual,
    title = "Continual Learning in Multilingual Sign Language Translation",
    author = "Yazdani, Shakib  and
      Genabith, Josef Van  and
      Espa{\~n}a-Bonet, Cristina",
    booktitle = "Proceedings of the 2025 Conference of the Nations of the Americas Chapter of the Association for Computational Linguistics: Human Language Technologies (Volume 1: Long Papers)",
    month = apr,
    year = "2025",
    address = "Albuquerque, New Mexico",
    publisher = "Association for Computational Linguistics",
    url = "https://aclanthology.org/2025.naacl-long.546/",
    doi = "10.18653/v1/2025.naacl-long.546",
    pages = "10923--10938",
}

@inproceedings{kocmi-federmann-2023-gemba,
    title = "{GEMBA}-{MQM}: Detecting Translation Quality Error Spans with {GPT}-4",
    author = "Kocmi, Tom  and
      Federmann, Christian",
    editor = "Koehn, Philipp  and
      Haddow, Barry  and
      Kocmi, Tom  and
      Monz, Christof",
    booktitle = "Proceedings of the Eighth Conference on Machine Translation",
    month = dec,
    year = "2023",
    address = "Singapore",
    publisher = "Association for Computational Linguistics",
    url = "https://aclanthology.org/2023.wmt-1.64/",
    doi = "10.18653/v1/2023.wmt-1.64",
    pages = "768--775"
}

@article{Tong_He_Shao_Yeung_2025, 
title={G-VEval: A Versatile Metric for Evaluating Image and Video Captions Using GPT-4o}, 
volume={39}, 
url={https://ojs.aaai.org/index.php/AAAI/article/view/32798}, 
DOI={10.1609/aaai.v39i7.32798}, 
abstractNote={Evaluation metric of visual captioning is important yet not thoroughly explored. Traditional metrics like BLEU, METEOR, CIDEr, and ROUGE often miss semantic depth, while trained metrics such as CLIP-Score, PAC-S, and Polos are limited in zero-shot scenarios. Advanced Language Model-based metrics also struggle with aligning to nuanced human preferences. To address these issues, we introduce G-VEval, a novel metric inspired by G-Eval and powered by the new GPT-4o. G-VEval uses chain-of-thought reasoning in large multimodal models and supports three modes: reference-free, reference-only, and combined, accommodating both video and image inputs. We also propose MSVD-Eval, a new dataset for video captioning evaluation, to establish a more transparent and consistent framework for both human experts and evaluation metrics. It is designed to address the lack of clear criteria in existing datasets by introducing distinct dimensions of Accuracy, Completeness, Conciseness, and Relevance (ACCR). Extensive results show that G-VEval outperforms existing methods in correlation with human annotations, as measured by Kendall tau-b and Kendall tau-c. This provides a flexible solution for diverse captioning tasks and suggests a straightforward yet effective approach for large language models to understand video content, paving the way for advancements in automated captioning.}, number={7}, 
journal={Proceedings of the AAAI Conference on Artificial Intelligence}, 
author={Tong, Tony Cheng and He, Sirui and Shao, Zhiwen and Yeung, Dit-Yan}, 
year={2025}, 
month={Apr.}, 
pages={7419-7427} 
}

@inproceedings{fomicheva-etal-2022-mlqe,
    title = "{MLQE}-{PE}: A Multilingual Quality Estimation and Post-Editing Dataset",
    author = "Fomicheva, Marina  and
      Sun, Shuo  and
      Fonseca, Erick  and
      Zerva, Chrysoula  and
      Blain, Fr{\'e}d{\'e}ric  and
      Chaudhary, Vishrav  and
      Guzm{\'a}n, Francisco  and
      Lopatina, Nina  and
      Specia, Lucia  and
      Martins, Andr{\'e} F. T.",
    editor = "Calzolari, Nicoletta  and
      B{\'e}chet, Fr{\'e}d{\'e}ric  and
      Blache, Philippe  and
      Choukri, Khalid  and
      Cieri, Christopher  and
      Declerck, Thierry  and
      Goggi, Sara  and
      Isahara, Hitoshi  and
      Maegaard, Bente  and
      Mariani, Joseph  and
      Mazo, H{\'e}l{\`e}ne  and
      Odijk, Jan  and
      Piperidis, Stelios",
    booktitle = "Proceedings of the Thirteenth Language Resources and Evaluation Conference",
    month = jun,
    year = "2022",
    address = "Marseille, France",
    publisher = "European Language Resources Association",
    url = "https://aclanthology.org/2022.lrec-1.530/",
    pages = "4963--4974"
}

@inproceedings{benkirane-etal-2024-machine,
    title = "Machine Translation Hallucination Detection for Low and High Resource Languages using Large Language Models",
    author = "Benkirane, Kenza  and
      Gongas, Laura  and
      Pelles, Shahar  and
      Fuchs, Naomi  and
      Darmon, Joshua  and
      Stenetorp, Pontus  and
      Adelani, David Ifeoluwa  and
      S{\'a}nchez, Eduardo",
    editor = "Al-Onaizan, Yaser  and
      Bansal, Mohit  and
      Chen, Yun-Nung",
    booktitle = "Findings of the Association for Computational Linguistics: EMNLP 2024",
    month = nov,
    year = "2024",
    address = "Miami, Florida, USA",
    publisher = "Association for Computational Linguistics",
    url = "https://aclanthology.org/2024.findings-emnlp.564/",
    doi = "10.18653/v1/2024.findings-emnlp.564",
    pages = "9647--9665"
}

@inproceedings{xu-etal-2024-pride,
    title = "Pride and Prejudice: {LLM} Amplifies Self-Bias in Self-Refinement",
    author = "Xu, Wenda  and
      Zhu, Guanglei  and
      Zhao, Xuandong  and
      Pan, Liangming  and
      Li, Lei  and
      Wang, William",
    editor = "Ku, Lun-Wei  and
      Martins, Andre  and
      Srikumar, Vivek",
    booktitle = "Proceedings of the 62nd Annual Meeting of the Association for Computational Linguistics (Volume 1: Long Papers)",
    month = aug,
    year = "2024",
    address = "Bangkok, Thailand",
    publisher = "Association for Computational Linguistics",
    url = "https://aclanthology.org/2024.acl-long.826/",
    doi = "10.18653/v1/2024.acl-long.826",
    pages = "15474--15492"
}

@inproceedings{papineni-etal-2002-bleu,
    title = "{B}leu: a Method for Automatic Evaluation of Machine Translation",
    author = "Papineni, Kishore  and
      Roukos, Salim  and
      Ward, Todd  and
      Zhu, Wei-Jing",
    editor = "Isabelle, Pierre  and
      Charniak, Eugene  and
      Lin, Dekang",
    booktitle = "Proceedings of the 40th Annual Meeting of the Association for Computational Linguistics",
    month = jul,
    year = "2002",
    address = "Philadelphia, Pennsylvania, USA",
    publisher = "Association for Computational Linguistics",
    url = "https://aclanthology.org/P02-1040/",
    doi = "10.3115/1073083.1073135",
    pages = "311--318"
}

@inproceedings{freitag-etal-2024-llms,
    title = "Are {LLM}s Breaking {MT} Metrics? Results of the {WMT}24 Metrics Shared Task",
    author = "Freitag, Markus  and
      Mathur, Nitika  and
      Deutsch, Daniel  and
      Lo, Chi-Kiu  and
      Avramidis, Eleftherios  and
      Rei, Ricardo  and
      Thompson, Brian  and
      Blain, Frederic  and
      Kocmi, Tom  and
      Wang, Jiayi  and
      Adelani, David Ifeoluwa  and
      Buchicchio, Marianna  and
      Zerva, Chrysoula  and
      Lavie, Alon",
    editor = "Haddow, Barry  and
      Kocmi, Tom  and
      Koehn, Philipp  and
      Monz, Christof",
    booktitle = "Proceedings of the Ninth Conference on Machine Translation",
    month = nov,
    year = "2024",
    address = "Miami, Florida, USA",
    publisher = "Association for Computational Linguistics",
    url = "https://aclanthology.org/2024.wmt-1.2/",
    doi = "10.18653/v1/2024.wmt-1.2",
    pages = "47--81"
}

@article{survey-of-hallucination,
author = {Ji, Ziwei and Lee, Nayeon and Frieske, Rita and Yu, Tiezheng and Su, Dan and Xu, Yan and Ishii, Etsuko and Bang, Ye Jin and Madotto, Andrea and Fung, Pascale},
title = {Survey of Hallucination in Natural Language Generation},
year = {2023},
issue_date = {December 2023},
publisher = {Association for Computing Machinery},
address = {New York, NY, USA},
volume = {55},
number = {12},
issn = {0360-0300},
url = {https://doi.org/10.1145/3571730},
doi = {10.1145/3571730},
journal = {ACM Comput. Surv.},
month = mar,
articleno = {248},
numpages = {38}
}

@article{guerreiro-etal-2023-hallucinations,
    title = "Hallucinations in Large Multilingual Translation Models",
    author = "Guerreiro, Nuno M.  and
      Alves, Duarte M.  and
      Waldendorf, Jonas  and
      Haddow, Barry  and
      Birch, Alexandra  and
      Colombo, Pierre  and
      Martins, Andr{\'e} F. T.",
    journal = "Transactions of the Association for Computational Linguistics",
    volume = "11",
    year = "2023",
    address = "Cambridge, MA",
    publisher = "MIT Press",
    url = "https://aclanthology.org/2023.tacl-1.85/",
    doi = "10.1162/tacl_a_00615",
    pages = "1500--1517"
}

@inproceedings{kim-etal-2024-signbleu,
    title = "{S}ign{BLEU}: Automatic Evaluation of Multi-channel Sign Language Translation",
    author = "Kim, Jung-Ho  and
      Huerta-Enochian, Mathew  and
      Ko, Changyong  and
      Lee, Du Hui",
    editor = "Calzolari, Nicoletta  and
      Kan, Min-Yen  and
      Hoste, Veronique  and
      Lenci, Alessandro  and
      Sakti, Sakriani  and
      Xue, Nianwen",
    booktitle = "Proceedings of the 2024 Joint International Conference on Computational Linguistics, Language Resources and Evaluation (LREC-COLING 2024)",
    month = may,
    year = "2024",
    address = "Torino, Italia",
    publisher = "ELRA and ICCL",
    url = "https://aclanthology.org/2024.lrec-main.1289/",
    pages = "14796--14811"
}

@InProceedings{imai-EtAl:2025:RANLP2,
  author    = {Imai, Saki  and  Inan, Mert  and  Sicilia, Anthony B.  and  Alikhani, Malihe},
  title     = {SiLVERScore: Semantically-Aware Embeddings for Sign Language Generation Evaluation},
  booktitle      = {Proceedings of the 15th International Conference on Recent Advances in Natural Language Processing - Natural Language Processing in the Generative AI era},
  month          = {September},
  year           = {2025},
  address        = {Varna, Bulgaria},
  publisher      = {INCOMA Ltd., Shoumen, Bulgaria},
  pages     = {452--461},
  url       = {https://aclanthology.org/2025.ranlp-1.54}
}

@article{Chen2022ASM,
  title={A Simple Multi-Modality Transfer Learning Baseline for Sign Language Translation},
  author={Yutong Chen and Fangyun Wei and Xiao Sun and Zhirong Wu and Stephen Lin},
  journal={2022 IEEE/CVF Conference on Computer Vision and Pattern Recognition (CVPR)},
  year={2022},
  pages={5110-5120},
  url={https://api.semanticscholar.org/CorpusID:247315595}
}

@article{kay2017kineticshumanactionvideo,
  title={The Kinetics Human Action Video Dataset},
  author={Will Kay and Jo{\~a}o Carreira and Karen Simonyan and Brian Zhang and Chloe Hillier and Sudheendra Vijayanarasimhan and Fabio Viola and Tim Green and Trevor Back and Apostol Natsev and Mustafa Suleyman and Andrew Zisserman},
  journal={ArXiv},
  year={2017},
  volume={abs/1705.06950},
  url={https://api.semanticscholar.org/CorpusID:27300853}
}

@inproceedings{li2020word,
      title={Word-level Deep Sign Language Recognition from Video: A New Large-scale Dataset and Methods Comparison},
      author={Li, Dongxu and Rodriguez, Cristian and Yu, Xin and Li, Hongdong},
      booktitle={The IEEE Winter Conference on Applications of Computer Vision},
      pages={1459--1469},
      year={2020}
}

@inproceedings{zeng-etal-2024-towards,
    title = "Towards Multiple References Era {--} Addressing Data Leakage and Limited Reference Diversity in Machine Translation Evaluation",
    author = "Zeng, Xianfeng  and
      Liu, Yijin  and
      Meng, Fandong  and
      Zhou, Jie",
    editor = "Ku, Lun-Wei  and
      Martins, Andre  and
      Srikumar, Vivek",
    booktitle = "Findings of the Association for Computational Linguistics: ACL 2024",
    month = aug,
    year = "2024",
    address = "Bangkok, Thailand",
    publisher = "Association for Computational Linguistics",
    url = "https://aclanthology.org/2024.findings-acl.710/",
    doi = "10.18653/v1/2024.findings-acl.710",
    pages = "11939--11951"
}

@inproceedings{mathur-etal-2020-tangled,
    title = "Tangled up in {BLEU}: Reevaluating the Evaluation of Automatic Machine Translation Evaluation Metrics",
    author = "Mathur, Nitika  and
      Baldwin, Timothy  and
      Cohn, Trevor",
    editor = "Jurafsky, Dan  and
      Chai, Joyce  and
      Schluter, Natalie  and
      Tetreault, Joel",
    booktitle = "Proceedings of the 58th Annual Meeting of the Association for Computational Linguistics",
    month = jul,
    year = "2020",
    address = "Online",
    publisher = "Association for Computational Linguistics",
    url = "https://aclanthology.org/2020.acl-main.448/",
    doi = "10.18653/v1/2020.acl-main.448",
    pages = "4984--4997"
}

@inproceedings{hamidullah-etal-2025-sonar,
    title = "{SONAR}-{SLT}: Multilingual Sign Language Translation via Language-Agnostic Sentence Embedding Supervision",
    author = "Hamidullah, Yasser  and
      Yazdani, Shakib  and
      Oguz, Cennet  and
      Van Genabith, Josef  and
      Espa{\~n}a-Bonet, Cristina",
    editor = "Haddow, Barry  and
      Kocmi, Tom  and
      Koehn, Philipp  and
      Monz, Christof",
    booktitle = "Proceedings of the Tenth Conference on Machine Translation",
    month = nov,
    year = "2025",
    address = "Suzhou, China",
    publisher = "Association for Computational Linguistics",
    url = "https://aclanthology.org/2025.wmt-1.18/",
    doi = "10.18653/v1/2025.wmt-1.18",
    pages = "301--313",
    ISBN = "979-8-89176-341-8"
}

@article{hamidullah2025grounding,
  title={Grounding or Guessing? Visual Signals for Detecting Hallucinations in Sign Language Translation},
  author={Hamidullah, Yasser and Chowdury, Koel Dutta and Al-Ghussin, Yusser and Yazdani, Shakib and Oguz, Cennet and van Genabith, Josef and Espa{\~n}a-Bonet, Cristina},
  journal={arXiv preprint arXiv:2510.18439},
  year={2025}
}

@article{likert1932technique,
  title={A technique for the measurement of attitudes},
  author={Likert, Rensis},
  journal={Archives of Psychology},
  volume={22},
  number={140},
  pages={1--55},
  year={1932}
}

\clearpage
\onecolumn

\appendix
\section{Prompts}

\subsection{Hallucination Detection}
\label{appendix:hallucination}

{\footnotesize
\begin{Verbatim}[fontsize=\small, commandchars=+\[\]]
System:

You are an expert evaluator of Sign Language Translation (SLT) outputs.

You will be given:
- A human reference translation
- A model predicted translation

Your task is to determine **how severe** the hallucination is in the
predicted translation.

Definition:
A word in the translated text is considered a hallucination if it introduces
information that is completely unrelated to the source text.

Assign one severity level according to these guidelines:

• No hallucination: The translated text does not contain any hallucinated words. 
• Small hallucination: The translated text contains 1-2 hallucinated words.  
• Partial hallucination: The translated text includes at least 3 
hallucinated words, but not all words are hallucinated.  
• Full hallucination: Nearly all words in the translated text are 
hallucinated, with the exception of perhaps 1-2 words.

Note: The labels are mutually exclusive; for example, a translation with
a partial hallucination does not qualify as a full hallucination.

User:
Reference Translation: {ref_text}
Predicted Translation: {mt_text}

Provide exactly one of the following labels as your response. Do not include any 
additional text or explanation:
• No hallucination
• Small hallucination
• Partial hallucination
• Full hallucination
\end{Verbatim}
}

\subsection{G-Eval}
\label{appendix:g-eval}
{\footnotesize
\begin{Verbatim}[fontsize=\small, commandchars=+\[\]]
You will be given a generated translation for a short sign language video segment, 
along with reference translation.

Your task is to rate the generated translation based on its Adequacy and Fluency
in capturing the intended meaning of the original signing as conveyed
in the reference translation.

Evaluation Criteria:

Score Range: 1 to 5 (integer) - The generated translation should:

- Correctly convey the meaning expressed in the reference translation.
- Include all key information without introducing unrelated or incorrect content.
- Be written in clear and natural language.
- Stay true to the meaning of the original signing.

Evaluation Dimensions:

1. Adequacy — Does the translation correctly convey the meaning?
2. Fluency — Is the translation grammatically correct, coherent, and easy to read?

Evaluation Steps:

1. Examine the reference translation to understand the 
meaning, key elements, and tone.
2. Read the generated translation thoroughly.
3. Compare the generated translation with the references and judge how well it:
    - Captures meaning accurately.
    - Covers all important elements.
    - Maintains grammatical fluency.
    - Stays faithful to the original intent.

4. Penalize for:
    - Missing key elements.
    - Introducing unrelated or incorrect details.
    - Awkward or unclear phrasing.
    - Changing the tone or meaning.

5. Assign an integer score from 1 to 5 for each dimension:
    - Adequacy score
    - Fluency score

Reference Translation:
{{Reference}}

Generated Translation:
{{Translation}}

Format of Output:
You should first give an explanation for each score, then end with 
two separate sentences:

..... The Adequacy score: {{Adequacy_score}}
..... The Fluency score: {{fluency_score}}

\end{Verbatim}
}

\subsection{GEMBA}
\label{appendix:gemba}
{\footnotesize
\begin{Verbatim}[fontsize=\small, commandchars=+\[\]]
Score the following translation from German Sign Language to German
with respect to the human reference, on a continuous scale from 0 to 100, 
where 0 means no meaning preserved and 100 means perfect meaning and grammar.

Scoring guidelines:

0–10: Incorrect translation (no relation to the source meaning).

11–29: Contains a few correct keywords, but the overall meaning 
differs significantly.

30–50: Major mistakes that distort meaning.

51–69: Understandable and conveys the main meaning but includes noticeable
grammatical or lexical errors.

70–90: Closely preserves the semantics of the source sentence,
with only minor issues.

91–100: Perfect translation—fully accurate, fluent, and natural.

Human reference: 
{{Reference}}
Model translation: 
{{Translation}}
..... Score: {{score}}
\end{Verbatim}
}

\end{document}